\documentclass[runningheads]{llncs}
\usepackage{graphicx}%
\usepackage{hyperref}
\usepackage{enumitem}
\usepackage{verbatim}
\usepackage{xcolor}
\usepackage{amssymb}
\usepackage{booktabs}
\usepackage{multirow} 
\usepackage{amssymb} %
\usepackage{tabularx}

\newcommand\samethanks[1][\value{footnote}]{\footnotemark[#1]}%
\makeatletter%
\def\@fnsymbol#1{\ensuremath{\ifcase#1\or\star\or\dagger\or\ddagger\or
   \mathchar "278\or \mathchar "27B\or \|\or **\or \dagger\dagger
   \or \ddagger\ddagger \else\@ctrerr\fi}}
\makeatother
\usepackage[T1]{fontenc}%

\usepackage{fancyvrb}
\definecolor{verbgray}{rgb}{0.3, 0.3, 0.3}
\DefineVerbatimEnvironment{myverbatim}{Verbatim}{fontfamily=cmtt,formatcom=\color{verbgray}}

\newcommand{\myparagraph}[1]{\noindent\textbf{#1.}\enspace}

\newcolumntype{Y}{>{\centering\arraybackslash}X}
\newcommand{\showstd}[1]{{\textcolor{gray}{#1}}}
\newcommand{\best}[1]{\textbf{#1}}
\newcommand{\bestci}[1]{\textbf{#1}}

\begin{document}
\title{Scalable Training of Spatially Grounded 2D Vision--Language Models for Radiology}
\titlerunning{Scalable Training of Spatially Grounded 2D VLMs for Radiology} %
\author{
Yusuf Salcan\inst{1,4}\thanks{Equal contribution.}
\and
Simon Ging\inst{1,2}\samethanks%
\and
Robin Tibor Schirrmeister\inst{3}
\and
Philipp Arnold\inst{3}%
\and
\\
Elmar Kotter\inst{3}%
\and
Behzad Bozorgtabar\inst{2}\thanks{Equal supervision.}%
\and
Thomas Brox\inst{1}\samethanks%
\\
\email{gings@cs.uni-freiburg.de}
\\
\texttt{\href{https://radgrounder.github.io}{radgrounder.github.io}}
}

\authorrunning{Y. Salcan, S. Ging et al.}
\institute{
Computer Vision Group, University of Freiburg, Germany
\and
Adaptive \& Agentic AI (A3) Lab, Aarhus University, Denmark
\and
Department of Radiology, Medical Center -- University of Freiburg, Germany
\and
CRIION-AI Lab, Freiburg, Germany
}
\maketitle

\begin{abstract}%
We study how to train visually grounded vision--language models (VLMs) for radiology without manual spatial annotations.
We introduce \textit{RefRad2D}, a large-scale bilingual (German/English) dataset of 1.2M CT and MRI image--text pairs derived from clinical practice, with task-specific VQA and spatial grounding subsets generated automatically via LLM-based curation and automated segmentation.
Trained on this data, our model \textit{RadGrounder} jointly performs report generation, visual question answering, and spatial grounding via bounding-box detection or segmentation.
On external VQA benchmarks (Slake, VQA-RAD), RadGrounder achieves competitive results with specialized medical VLMs.
Adding our clinical data to the training mixture improves open-ended VQA over fine-tuning on the downstream datasets alone, showing the transferability of our dataset.
Crucially, adding grounding supervision does not degrade language quality, enabling spatially verifiable outputs at no cost to VQA performance.
\keywords{Radiology VLM \and Visual grounding \and CT \and MRI}
\end{abstract}
\section{Introduction}
Vision--language models (VLMs) for radiology can generate coherent medical text, but a significant limitation remains: they cannot reliably ground their output in specific image regions.
Without spatial grounding, predictions are difficult to verify, raising concerns about ``hallucinations'' and limiting clinical trust.
Training such models for CT and MRI is further hindered by data scarcity.
Approaches that consume 2D slices offer a scalable alternative to 3D volumetric methods, yet prior efforts have struggled with scale; for instance, Med-Gemini authors \cite{medgemini1,medgemini2} tailored a 2D CT slice dataset but their strict filtering yielded only 4,009 images, leaving 2D slice supervision at scale unexplored.

Recent medical VLMs achieve strong performance on radiology benchmarks, from lightweight generalists like BiomedGPT~\cite{biomedgpt} and LLaVA-Med~\cite{Li2024_llavamed} to large-scale models like RadFM~\cite{wu2023towards}, which supports both 2D and 3D inputs.
Most work on spatially grounded medical VLMs has focused on chest X-rays, including CheXagent \cite{chen2024chexagent} and MAIRA-2 \cite{maira2}.
VividMed \cite{vividmed} extended grounding to broader medical imaging, but relies on synthetic data from open datasets.
Grounded VLMs trained on large-scale clinical CT and MRI data remain underexplored.

In this work, we introduce RadGrounder, a PaliGemma~2-based \cite{PaliGemmaRef2024} multi-task VLM that jointly performs report generation, VQA, and visual grounding on CT and MRI slices.
To train it, we build RefRad2D, a bilingual (German/English) corpus of 1.2M image--text pairs derived from clinical routine via an automated LLM-driven curation pipeline, with spatial labels from TotalSegmentator~\cite{totalsegmentator_ct,totalsegmentator_mr}.
\\\\
\myparagraph{Contributions:}
\begin{itemize}[leftmargin=*,itemsep=2pt,topsep=2pt]
    \item We present RefRad2D, a large-scale bilingual (German/English) 2D CT/MRI dataset for VLM training with 1.2M image--text pairs and automatically derived spatial grounding annotations.
    \item We introduce RadGrounder, a multi-task architecture that jointly performs VQA and visual grounding, demonstrating that 2D slice-level supervision derived from routine reports is a scalable strategy for training radiology VLMs. Code and pretrained models will be made publicly available upon acceptance.
    \item We ablate training configurations and grounding strategies, finding that (1) token-based bounding box prediction provides effective localization without requiring an auxiliary segmentation head or additional loss terms, and (2) adding spatial grounding supervision does not degrade VQA or report generation quality.
    \item We validate our approach on external benchmarks (VQA-RAD~\cite{vqarad}, Slake~\cite{slake}), showing competitive performance with other medical VLMs while enabling verifiable, spatially grounded interpretations.
\end{itemize}

\section{Dataset}
\label{sec:dataset}

We introduce RefRad2D, a bilingual dataset with visual grounding annotations, derived from clinical logs at a university hospital. The dataset comprises 1.2 million unique image--caption pairs extracted from 945k CT and 321k MRI slices, representing a decade of clinical practice \cite{smithbindman2019trends}.

\myparagraph{Preprocessing and Bilingual Expansion} Raw reports often contain longitudinal references (e.g., ``lesion has grown'') that induce hallucinations in single-frame VLMs. We used GPT-OSS~(120B)~\cite{gptoss} to rewrite captions, removing temporal context while preserving current findings \cite{thirunavukarasu2023llms}. To address Anglo-centric bias \cite{neveol2018clinical}, we translated all German reports into English using Gemma~3~(27B)~\cite{gemma3}. To refine translation quality, we sampled outputs and judged them with a stronger model, GPT-OSS~(120B), iteratively improving the translation prompt based on identified errors~\cite{dubois2024alpacafarm}. We train on a mix of both languages.

\myparagraph{Automated Anatomical Grounding}
To generate dense, pixel-level annotations without manual labeling, we developed an automated pipeline using 181,362 CT and 36,026 MRI 3D volumes. This produced the RefRad2D-Grounded subset: 236,157 grounded slice--text pairs (217,692 CT, 18,465 MRI).

\myparagraph{Anatomical Segmentation} We processed full 3D volumes using TotalSegmentator~\cite{totalsegmentator_ct,totalsegmentator_mr} to generate masks for 117 CT and 50 MRI classes. These were harmonized into a unified schema of $C=121$ classes by merging shared anatomies. Volumes were sliced to match the VLM's 2D input format.

\begin{figure}[t]
    \centering
    \includegraphics[width=\textwidth]{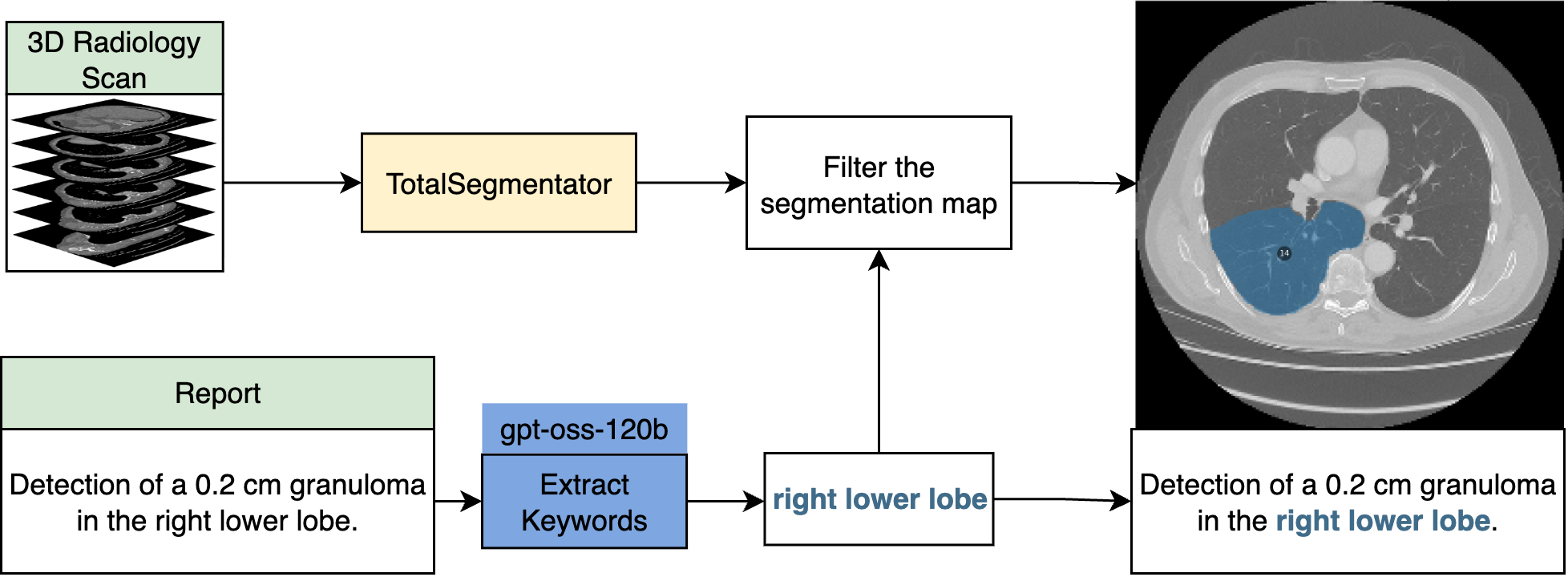} 
    \caption{Overview of the components to generate the data for visual grounding. We use TotalSegmentator~\cite{totalsegmentator_ct,totalsegmentator_mr} to create 3D segmentation masks. Via slicing and LLM-based keyword extraction we create dense 2D pixel-level annotations. A strict set-intersection filter ($C_T \cap C_I \neq \emptyset$) ensures high-quality matching between textual entities and visual regions.}
    \label{fig:grounding_pipeline}
\end{figure}

\myparagraph{Keyword-Mask Matching} To link visual regions to text (Fig.~\ref{fig:grounding_pipeline}), we used GPT-OSS~(120B) to extract anatomical mentions from each caption and map them to the same $C=121$ class schema used by TotalSegmentator. Let $C_T \subseteq \{1,\ldots,C\}$ denote the classes mentioned in caption $T$, $C_I$ the classes with masks in slice $I$, and $M_I^T$ the segmentation mask for this class in this slice. We construct a valid training triplet $(I, T, M_I^T)$ only if $C_T \cap C_I \neq \emptyset$.
For detection, we convert the masks into bounding boxes by taking the box that covers the mask.

\myparagraph{Synthetic VQA Generation}
For VQA training, we generated the RefRad2D-VQA dataset ($\sim$9.6M pairs). Using Gemma~3~(27B), we derived 5 QA pairs per image directly from clinical findings (Open, Yes/No, Multiple Choice) and 3 QA pairs from the slice metadata \cite{Li2024_llavamed}.

\section{Method}
\label{sec:method}

We propose RadGrounder, a multi-task Vision--Language Model (VLM) that generates radiology reports and localizes findings through bounding-box detection or segmentation.

\myparagraph{Architecture}
Our architecture builds upon the PaliGemma~2~(3B) framework \cite{PaliGemmaRef2024}, comprising a SigLIP-So400m~\cite{siglip} vision encoder and a Gemma-2B language decoder~\cite{Gemma2024}. The model processes an image $I \in \mathbb{R}^{H \times W \times 3}$ into visual tokens $Z_v$, which are concatenated with text tokens $Z_t$ and are processed autoregressively. We investigate two grounding strategies on top of this foundation: token-based bounding-box detection, which the model can perform natively via text generation, and an auxiliary segmentation head.

\myparagraph{Bounding-Box Detection} For spatial grounding, we treat detection as a text-generation task. We extend the vocabulary with coordinate tokens discretized into 512 bins and class-identifying tokens. The model generates a structured sequence:
\begin{equation}
    \texttt{<p bbox> [LOC] id=<segID> KEYWORD </p>}
\end{equation}
where \texttt{[LOC]} denotes the bounding box coordinates $\langle y_{min}, x_{min}, y_{max}, x_{max} \rangle$ and \texttt{id} maps to our unified schema of $C=121$ anatomical classes. To resolve ambiguity in slices containing multiple instances of the same organ (e.g., multiple lymph nodes), we employ a class-wise merging strategy, predicting a single union bounding box for all instances of a class.

\myparagraph{Segmentation Head} We also explore pixel-level grounding via a lightweight mask decoder following VividMed~\cite{vividmed} and SAM~\cite{sam}. When the model generates a \texttt{</seg>} token, its hidden state is projected into a 256-dim prompt that drives a Two-Way Transformer~\cite{sam} over the image embeddings to produce a binary mask. Unlike detection, this approach requires an auxiliary decoder and an additional segmentation loss.

\myparagraph{Training Objective}
The model is trained end-to-end with auto-regressive cross-entropy $\mathcal{L}_{txt}$ for next-token prediction. Detection grounding is trained entirely through $\mathcal{L}_{txt}$, as bounding-box coordinates are generated as tokens. For the segmentation variant, we add an auxiliary loss:
\begin{equation}
    \mathcal{L} = \mathcal{L}_{txt} + \lambda_{seg} (\mathcal{L}_{focal} + \mathcal{L}_{dice})
\end{equation}
where $\mathcal{L}_{focal}$ and $\mathcal{L}_{dice}$ handle the class imbalance inherent in medical segmentation. We employ a dynamic sampling strategy where batches alternate between report generation, VQA, and grounded tasks.

\myparagraph{LLMScore}
\label{subsec:metrics}
Standard n-gram metrics (e.g., CIDEr) struggle with semantic equivalence and the complexities of medical text. We developed LLMScore by adopting the ``LLM-as-a-judge'' paradigm~\cite{zheng2023judging,ovqa}. We use Gemma~3~(27B) as the evaluator to compare the generated report against the ground truth. The judge assesses clinical factuality and semantic correctness on a 5-point scale normalized to $[0,1]$ and outputs a brief textual justification for its decision. To validate this metric, three radiologists independently scored 200 QA pairs from VQA-RAD. Inter-annotator agreement was high (ordinal Krippendorff's $\alpha = 0.958$)~\cite{krippendorff}, and LLMScore achieved a Pearson correlation of $r = 0.977$ with the mean human scores, indicating strong alignment with expert judgment.

\providecolor{ioBg}{RGB}{245,245,245}
\providecolor{metricsBg}{RGB}{235,245,255}
\begin{figure}
\centering
\begin{minipage}[t]{0.34\linewidth}
    \vspace{0pt}
    \includegraphics[width=\linewidth]{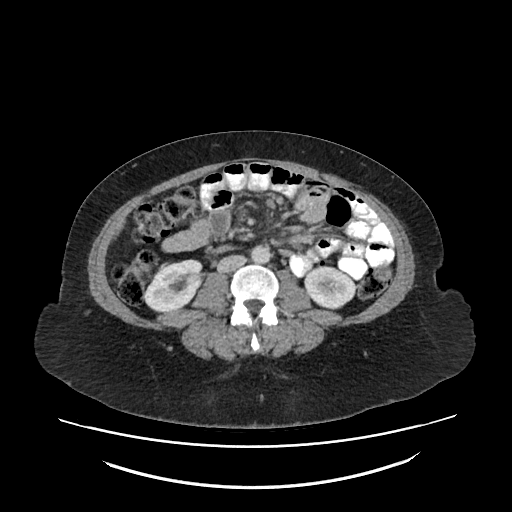}
\end{minipage}\hfill
\begin{minipage}[t]{0.64\linewidth}
    \vspace{0pt}
    \raggedright
    \setlength{\fboxsep}{2pt}
    \colorbox{ioBg}{\parbox{\dimexpr\linewidth-2\fboxsep\relax}{%
    \textbf{Prompt:} <image>
Clinical information: Status post sleeve gastrectomy, bypass, with increasing abdominal pain.
Question: Internal hernia?
Caption this image:\par
    \textbf{Ground Truth:} Nonspecific increased mesenteric lymph nodes, up to 10mm.\par
    \textbf{Prediction:} Increased mesenteric lymph nodes up to 1.2 cm.
    }}\par
    \vspace{1pt}
    \setlength{\fboxsep}{2pt}
    \colorbox{metricsBg}{\parbox{\dimexpr\linewidth-2\fboxsep\relax}{%
    \textbf{Metrics:}\par \textbf{CIDEr}: 1.0877\par \textbf{LLM Score}: 0.7500\par \textbf{LLM Reason}: The candidate and reference both identify increased mesenteric lymph nodes. The sizes are very close (1.2cm vs 10mm = 1cm), indicating a mostly right answer.
    }}
\end{minipage}
\caption{Qualitative example of RadGrounder's report generation on an abdominal CT slice (soft tissue window). The model successfully identifies increased mesenteric lymph nodes and accurately estimates their size. Despite a low traditional CIDEr score (1.08), the LLM-as-a-judge correctly assigns a high factual score (0.75) by recognizing the semantic equivalence of the measurements (1.2 cm vs. 10 mm).}
\label{fig:figure_examples_v1_01}
\end{figure}

\providecolor{ioBg}{RGB}{245,245,245}
\providecolor{metricsBg}{RGB}{235,245,255}
\providecolor{predTextColor}{RGB}{200,0,0}
\providecolor{gtTextColor}{RGB}{0,140,0}
\begin{figure}
\centering
\begin{minipage}[t]{0.34\linewidth}
    \vspace{0pt}
    \includegraphics[width=\linewidth]{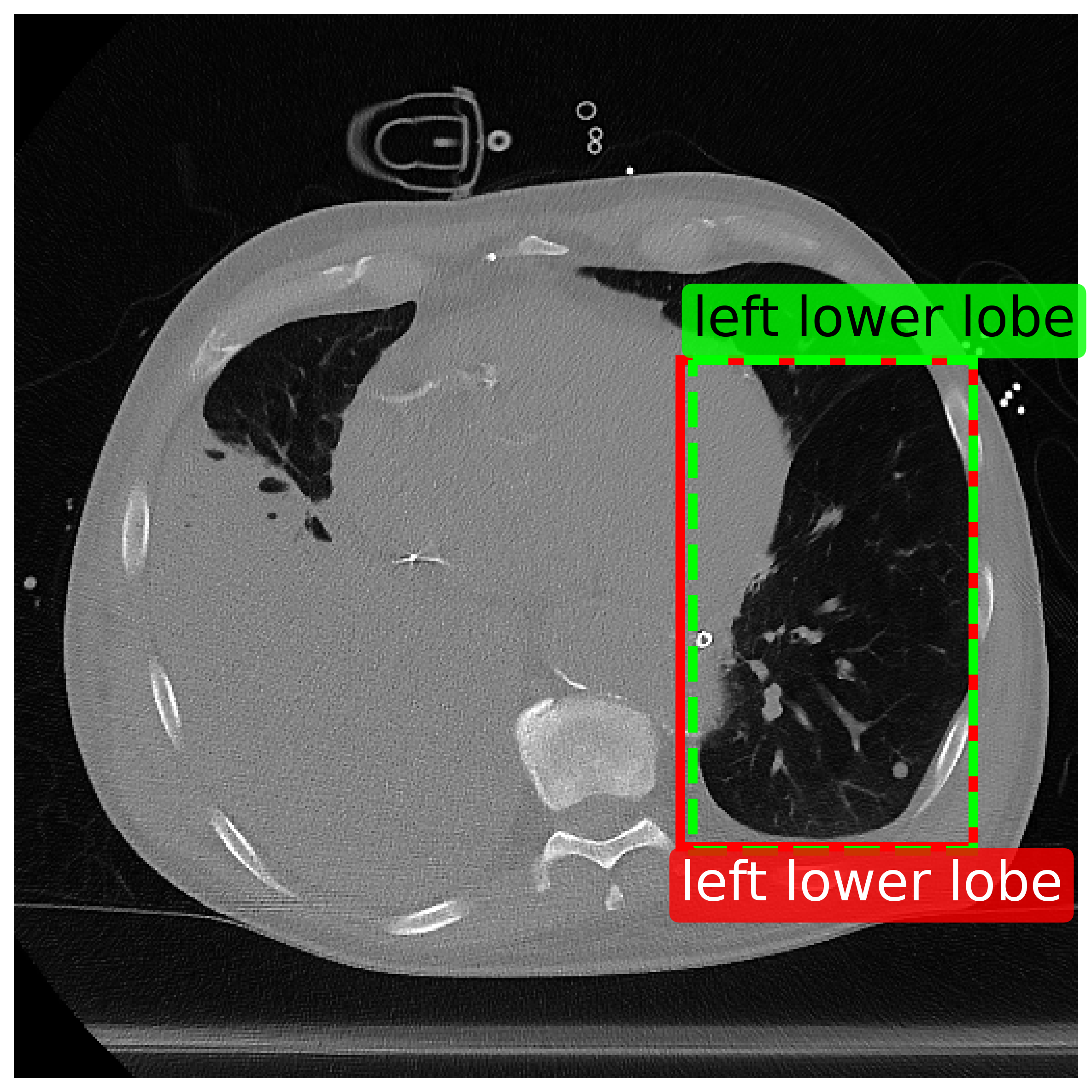}
\end{minipage}\hfill
\begin{minipage}[t]{0.64\linewidth}
    \vspace{0pt}
    \raggedright
    \setlength{\fboxsep}{2pt}
    \colorbox{ioBg}{\parbox{\dimexpr\linewidth-2\fboxsep\relax}{%
    \textbf{Prompt:} <image> Clinical information: Severe pneumonia in a patient with COPD Gold 4. Question: Extent of the pneumonia? Caption this image and detect keywords:\par
    \textbf{Ground Truth:} Approximately 6 mm measuring round lesion in the \textcolor{gtTextColor}{\textbf{left lower lobe}}.\par
    \textbf{Prediction:} Nodular densities in the left lung, exemplified in the \textcolor{predTextColor}{\textbf{left lower lobe}} with 5 mm, most likely postinflammatory.
    }}\par
    \vspace{1pt}
    \setlength{\fboxsep}{2pt}
    \colorbox{metricsBg}{\parbox{\dimexpr\linewidth-2\fboxsep\relax}{%
    \textbf{Metrics:}\par \textbf{CIDEr}: 0.1373\par \textbf{LLM Score}: 0.7500\par \textbf{LLM Reason}: The candidate and reference both identify a lesion in the left lower lobe and provide similar size estimations (5mm vs 6mm). The candidate adds 'nodular densities' and 'postinflammatory' which aren't explicitly in the reference but are plausible given the clinical context.\par \textbf{G-IoU}: 0.9497\par \textbf{mIoU}: 0.9497
    }}
\end{minipage}
\caption{Detection grounding performance on a chest CT (lung window). RadGrounder accurately localizes a nodular lesion in the left lower lobe, generating a bounding box that strongly aligns spatially and semantically with the ground truth (G-IoU: 0.9497).}
\label{fig:figure_detection_01}
\end{figure}

\myparagraph{Spatial-Semantic Evaluation (G-IoU)} Standard Intersection over Union (IoU) assumes a fixed vocabulary and fails when a model generates a keyword that is semantically correct but lexically distinct from the ground truth. We propose Grounding-IoU (G-IoU) to jointly measure spatial and semantic fidelity.
We first match predicted and ground-truth entities using the cosine similarity of their text embeddings. The G-IoU is then computed as the sum of their spatial IoUs weighted by this semantic similarity. The score is normalized by the maximum number of predicted or ground-truth entities, penalizing both spatial hallucinations (false positives) and diagnostic omissions (false negatives).

\begin{table*}
{
\centering
\setlength{\tabcolsep}{2pt}
\caption{Quantitative results on the internal RefRad2D test set for Report Captioning, VQA, and spatial grounding tasks. The table compares our RadGrounder detection model (row 4) against PaliGemma~2~\cite{PaliGemmaRef2024} (PaliG.~2), Gemma~3~\cite{gemma3}, and MedGemma~\cite{medgemma} (MedG.) (top), and ablates various training configurations (bottom). SL.: SigLIP vision encoder. We highlight \textbf{values within the confidence interval of the highest score} separately for the baseline comparison and the ablations.}
\label{tab:freirad_dataset_performance}
\begin{tabular}{@{}llllll|rcrc|rcrcrc|rc@{}}
\toprule
\multicolumn{6}{l|}{Training data} & \multicolumn{4}{c|}{Report} & \multicolumn{6}{c|}{VQA} & \multicolumn{2}{c}{Grounded} \\
\midrule
\multirow{2}{*}{\rotatebox{90}{\hspace{0.7em}Report}} & \multirow{2}{*}{\rotatebox{90}{\hspace{0.7em}VQA}} & \multirow{2}{*}{\rotatebox{90}{\hspace{0.7em}Slake}} & \multirow{2}{*}{\rotatebox{90}{\hspace{0.7em}VQA-RAD}} & \multirow{2}{*}{\rotatebox{90}{\hspace{0.7em}Detect.}} & \multirow{2}{*}{\rotatebox{90}{\hspace{0.7em}Segment}} & \multicolumn{2}{c}{CIDEr} & \multicolumn{2}{c|}{LLM} & \multicolumn{2}{c}{F1} & \multicolumn{2}{c}{LLM} & \multicolumn{2}{c|}{Acc.} & \multicolumn{2}{c}{G-IoU} \\
 &  &  &  &  &  & \multicolumn{2}{c}{} & \multicolumn{2}{c|}{} & \multicolumn{2}{c}{Open} & \multicolumn{2}{c}{Open} & \multicolumn{2}{c|}{Closed} & \multicolumn{2}{c}{} \\[6ex]
\midrule
\multicolumn{6}{l|}{PaliG. 2 (3B)~\cite{PaliGemmaRef2024}} & 0.6 & \showstd{0.1} & 2.6 & \showstd{0.5} & 12.7 & \showstd{1.9} & 22.8 & \showstd{2.3} & 51.5 & \showstd{2.7} & -- &  \\
\multicolumn{6}{l|}{Gemma 3 (4B)~\cite{gemma3}} & 0.1 & \showstd{0.0}& 21.7 & \showstd{1.3} & 5.5 & \showstd{0.8} & 31.8 & \showstd{2.7} & 13.0 & \showstd{1.6} & -- &  \\
\multicolumn{6}{l|}{MedG. (4B)~\cite{medgemma}} & 0.4 & \showstd{0.1} & 21.8 & \showstd{1.3} & 9.0 & \showstd{1.2} & 40.6 & \showstd{2.9} & 16.6 & \showstd{2.2} & -- &  \\
\midrule
\midrule
$\checkmark$ & $\checkmark$ & $\checkmark$ & $\checkmark$ & $\checkmark$ &  & \best{37.5} & \showstd{3.3} & \best{30.3} & \showstd{1.4} & \best{48.9} & \showstd{2.8} & \best{60.1} & \showstd{2.6} & \best{94.6} & \showstd{1.4} & \best{43.6} & \showstd{1.6} \\
\midrule
\multicolumn{6}{l|}{+ Unfreeze finet. SL.} & 34.6 & \showstd{3.3} & 28.0 & \showstd{1.4} & \bestci{47.8} & \showstd{2.8} & 60.1 & \showstd{2.5} & \bestci{95.1} & \showstd{1.5} & 43.4 & \showstd{1.7} \\
\multicolumn{6}{l|}{+ Unfreeze orig. SL.} & \bestci{36.9} & \showstd{3.2} & \bestci{29.9} & \showstd{1.4} & \bestci{49.2} & \showstd{2.8} & 60.8 & \showstd{2.6} & 94.0 & \showstd{1.5} & \best{45.2} & \showstd{1.7} \\
\midrule
$\checkmark$ & $\checkmark$ & $\checkmark$ & $\checkmark$ &  & $\checkmark$ & \bestci{36.9} & \showstd{3.2} & \bestci{30.1} & \showstd{1.4} & \best{50.4} & \showstd{2.8} & \best{64.1} & \showstd{2.7} & \best{95.6} & \showstd{1.3} & 36.9 & \showstd{1.6} \\
$\checkmark$ & $\checkmark$ &  &  &  &  & \bestci{39.0} & \showstd{3.4} & \best{30.7} & \showstd{1.5} & \bestci{49.2} & \showstd{2.8} & 60.3 & \showstd{2.6} & \bestci{94.5} & \showstd{1.5} & -- &  \\
 &  & $\checkmark$ & $\checkmark$ &  &  & 0.1 & \showstd{0.0} & 1.5 & \showstd{0.5} & 15.7 & \showstd{2.2} & 21.8 & \showstd{2.5} & 32.4 & \showstd{2.2} & -- &  \\
$\checkmark$ & $\checkmark$ & $\checkmark$ & $\checkmark$ &  &  & \bestci{37.3} & \showstd{3.3} & \bestci{30.6} & \showstd{1.4} & \bestci{50.1} & \showstd{2.8} & 61.2 & \showstd{2.6} & \bestci{94.4} & \showstd{1.5} & -- &  \\
$\checkmark$ & $\checkmark$ &  &  & $\checkmark$ &  & \best{39.2} & \showstd{3.3} & \bestci{30.6} & \showstd{1.5} & \bestci{49.1} & \showstd{2.8} & 60.2 & \showstd{2.6} & \bestci{94.5} & \showstd{1.5} & \bestci{43.6} & \showstd{1.6} \\
\bottomrule
\end{tabular}}
\end{table*}

\section{Experiments}
\label{sec:experiments}

\begin{table*}[t]
{
\centering
\setlength{\tabcolsep}{1.8pt}
\caption{Performance on Slake~\cite{slake} and VQA-RAD~\cite{vqarad}. SL.: SigLIP vision encoder, PaliG.: PaliGemma, MedG.: MedGemma. We used the train/test split from Xu et al.~\cite{xu2023elixrgeneralpurposexray} due to overlap between train and test images in the original VQA-RAD splits~\cite{medgemini2}. *Reported results are not available for that split.
We highlight \textbf{values within the confidence interval of the highest score} separately for the baseline comparison and the ablations.}
\label{tab:other_vqa_performance}
\begin{tabular}{@{}llllll|rcrcrcrc|rcrcrc@{}}
\toprule
\multicolumn{6}{l|}{Training data} & \multicolumn{8}{c|}{Slake} & \multicolumn{6}{c}{VQA-RAD} \\
\midrule
\multirow{2}{*}{\rotatebox{90}{\hspace{0.7em}Report}} & \multirow{2}{*}{\rotatebox{90}{\hspace{0.7em}VQA}} & \multirow{2}{*}{\rotatebox{90}{\hspace{0.7em}Slake}} & \multirow{2}{*}{\rotatebox{90}{\hspace{0.7em}VQA-RAD}} & \multirow{2}{*}{\rotatebox{90}{\hspace{0.7em}Detect.}} & \multirow{2}{*}{\rotatebox{90}{\hspace{0.7em}Segment}} & \multicolumn{2}{c}{F1} & \multicolumn{2}{c}{Recall} & \multicolumn{2}{c}{LLM} & \multicolumn{2}{c|}{Acc.} & \multicolumn{2}{c}{F1} & \multicolumn{2}{c}{LLM} & \multicolumn{2}{c}{Acc.} \\
 &  &  &  &  &  & \multicolumn{2}{c}{} & \multicolumn{2}{c}{Open} & \multicolumn{2}{c}{Open} & \multicolumn{2}{c|}{Closed} & \multicolumn{2}{c}{} & \multicolumn{2}{c}{Open} & \multicolumn{2}{c}{Closed} \\[6ex]
\midrule
\multicolumn{6}{l|}{PaliG. 2 (3B)~\cite{PaliGemmaRef2024} } & 24.5 &  & 24.3 &  & 34.2 & \showstd{3.5} & 58.1 &  & 37.9 &  & 26.0 & \showstd{4.1} & 55.6 &  \\
\multicolumn{6}{l|}{Gemma 3 (4B)~\cite{gemma3}} & 40.2 &  & 33.3 &  & 22.3 & \showstd{3.2} & 53.0 &  & 33.6 &  & 11.1 & \showstd{3.5} & 33.6 &  \\
\multicolumn{6}{l|}{MedG. (4B)~\cite{medgemma}} & 72.3 &  & 63.3 &  & 54.9 & \showstd{3.8} & \bestci{87.6} &  & \bestci{49.9} &  & 20.8 & \showstd{4.2} & 69.1 &  \\
\multicolumn{6}{l|}{BiomedGPT-B~\cite{biomedgpt}} & 85.2 &  & -- &  & -- &  & \bestci{89.9} &  & \multicolumn{2}{c}{*} & -- &  & \multicolumn{2}{c}{*} \\
\multicolumn{6}{l|}{LLaVA-Med~\cite{Li2024_llavamed}} & -- &  & \bestci{87.1} &  & -- &  & 86.8 &  & \multicolumn{2}{c}{*} & -- &  & \multicolumn{2}{c}{*} \\
\multicolumn{6}{l|}{RadFM (14B)~\cite{wu2023towards}} & 84.4 &  & -- &  & -- &  & \multicolumn{2}{c|}{--} & \multicolumn{2}{c}{*} & -- &  & -- &  \\
\multicolumn{6}{l|}{Med-Gemini~\cite{medgemini2}} & 75.8 &  & -- &  & -- &  & 84.8 &  & \multicolumn{2}{c}{*} & -- &  & \best{78.8} &  \\
\midrule
\midrule
$\checkmark$ & $\checkmark$ & $\checkmark$ & $\checkmark$ & $\checkmark$ &  & \best{87.7} & \showstd{2.4} & \best{88.1} & \showstd{2.4} & \best{88.8} & \showstd{2.3} & \best{90.3} & \showstd{2.7} & \best{50.7} & \showstd{3.5} & \best{44.3} & \showstd{4.8} & 64.7 & \showstd{4.6} \\
\midrule
\multicolumn{6}{l|}{Unfreeze finet. SL.} & \best{88.4} & \showstd{2.4} & \best{88.9} & \showstd{2.4} & \best{89.9} & \showstd{2.3} & \bestci{90.8} & \showstd{2.6} & \bestci{48.7} & \showstd{3.4} & \bestci{44.3} & \showstd{4.8} & 62.7 & \showstd{4.6} \\
\multicolumn{6}{l|}{Unfreeze orig. SL.} & \bestci{87.6} & \showstd{2.3} & \bestci{88.2} & \showstd{2.3} & \bestci{88.6} & \showstd{2.2} & \best{91.7} & \showstd{2.7} & \bestci{50.2} & \showstd{3.4} & \bestci{44.9} & \showstd{4.8} & \bestci{64.5} & \showstd{4.6} \\
\midrule
$\checkmark$ & $\checkmark$ & $\checkmark$ & $\checkmark$ &  & $\checkmark$ & \bestci{86.4} & \showstd{2.5} & \bestci{86.9} & \showstd{2.5} & 87.3 & \showstd{2.5} & \bestci{89.6} & \showstd{2.8} & \best{50.4} & \showstd{3.5} & \best{46.6} & \showstd{5.2} & \bestci{65.9} & \showstd{4.4} \\
$\checkmark$ & $\checkmark$ &  &  &  &  & 19.2 & \showstd{2.9} & 20.3 & \showstd{3.0} & 31.9 & \showstd{3.4} & 63.0 & \showstd{4.5} & 39.6 & \showstd{3.4} & 33.0 & \showstd{4.5} & 52.9 & \showstd{4.7} \\
 &  & $\checkmark$ & $\checkmark$ &  &  & \bestci{86.9} & \showstd{2.5} & \bestci{87.6} & \showstd{2.5} & \bestci{87.8} & \showstd{2.4} & 88.4 & \showstd{3.2} & \bestci{47.9} & \showstd{3.4} & 28.6 & \showstd{4.2} & \best{68.1} & \showstd{4.4} \\
$\checkmark$ & $\checkmark$ & $\checkmark$ & $\checkmark$ &  &  & \bestci{87.2} & \showstd{2.5} & \bestci{87.6} & \showstd{2.4} & \bestci{88.8} & \showstd{2.3} & \bestci{90.8} & \showstd{2.7} & \bestci{49.2} & \showstd{3.4} & \bestci{42.8} & \showstd{4.7} & 63.2 & \showstd{4.5} \\
$\checkmark$ & $\checkmark$ &  &  & $\checkmark$ &  & 18.3 & \showstd{2.8} & 20.2 & \showstd{3.0} & 31.4 & \showstd{3.4} & 58.8 & \showstd{4.7} & 40.2 & \showstd{3.4} & 35.9 & \showstd{4.5} & 52.9 & \showstd{4.9} \\
\bottomrule
\end{tabular}}
\end{table*}

\myparagraph{Implementation Details}
All models were trained on a single NVIDIA H100 GPU for 6 epochs with a batch size of 24 and 12 gradient accumulation steps, using the Adafactor optimizer with a learning rate of 5e-5. We apply random scale cropping, contrast adjustment, and intensity shifts as data augmentation. Training with the frozen SigLIP encoder on the full dataset mixture took approximately 2.5 days.

\myparagraph{Datasets and Protocols}
The RefRad2D training split contains 760,409 CT and 256,197 MRI image--report pairs. The RefRad2D-VQA subset provides 8 QA pairs per image, and the RefRad2D-Grounded training subset contains 174,655 CT and 14,721 MRI spatially grounded triplets. All internal evaluation sets were standardized to 2,000 test samples. For external benchmarking, we use Slake~\cite{slake} (9,849 training / 2,070 testing QA pairs) and VQA-RAD~\cite{vqarad} (3,064 training / 451 testing QA pairs).
For open-ended evaluation we use LLMScore (Sec.~\ref{subsec:metrics}).
We report 95\% bootstrap confidence intervals ($B{=}10{,}000$) over test set samples.

\myparagraph{Vision Encoder Strategy (Tab.~\ref{tab:freirad_dataset_performance},~\ref{tab:other_vqa_performance})}
Before multi-task training, we adapt the SigLIP encoder to the medical domain via contrastive pretraining on RefRad2D. We then ablate whether to keep this adapted encoder frozen or unfrozen during the subsequent LLM training, and compare against unfreezing the original (non-adapted) SigLIP weights. All three strategies perform comparably within confidence intervals on both internal and external benchmarks. Since freezing the encoder reduces memory footprint and accelerates training, we adopt it as our standard configuration.

\myparagraph{Dataset Mixture and Grounding (Tab.~\ref{tab:freirad_dataset_performance},~\ref{tab:other_vqa_performance})}
Baseline models score near zero on report captioning CIDEr, though LLMScore is less affected (e.g., MedGemma: 0.4 vs.\ 21.8). CIDEr penalizes any n-gram mismatch with the reference, while LLMScore captures semantic correctness (Fig.~\ref{fig:figure_examples_v1_01}).
The low baselines are expected: these models were not trained on our data, and the task requires generating hospital-specific text from a single slice without clinical context.
Token-based detection (Fig.~\ref{fig:figure_detection_01}) achieved higher G-IoU (43.6 vs.\ 36.9) than the segmentation head while requiring no additional loss terms or decoder parameters.
Neither grounding strategy degrades VQA or report generation quality compared to the no-grounding baseline. Mixture ablations show clear domain effects: training exclusively on internal data achieved the highest report generation score (CIDEr: 39.2) but failed to generalize to external VQA (Slake F1: 18.3). Conversely, adding our internal corpus to external fine-tuning improved open-ended VQA on most metrics (e.g., Slake F1: 87.7 vs.\ 86.9; VQA-RAD F1: 50.7 vs.\ 47.9). The exception was VQA-RAD closed accuracy, where the external-only model scored higher (68.1\% vs.\ 64.7\%).

\myparagraph{Comparison to State-of-the-Art (Tab.~\ref{tab:other_vqa_performance})}
Using our standard configuration (frozen encoder, full multi-task mixture), RadGrounder achieves competitive performance on external benchmarks. On Slake, our model reaches F1 87.7 and Closed Accuracy 90.3, comparable to BiomedGPT-B~\cite{biomedgpt} (85.2 / 89.9) and LLaVA-Med~\cite{Li2024_llavamed} (86.8 Acc), while surpassing Med-Gemini~\cite{medgemini1,medgemini2} (75.8 / 84.8) and MedGemma~\cite{medgemma} (72.3 / 87.6). On VQA-RAD, our model achieves the highest Open F1 (50.7) among compared methods.

\section{Conclusion}
We presented RadGrounder, a visually grounded VLM for radiology trained without manual spatial annotations, and RefRad2D, a bilingual dataset of 1.2M CT and MRI image--text pairs with automatically derived spatial labels. On external VQA benchmarks, RadGrounder achieves competitive performance with specialized medical VLMs, and adding our clinical data improves over fine-tuning on downstream data alone. We explored two grounding strategies---token-based detection and auxiliary segmentation---finding that detection provides effective localization without additional model components or loss terms, and that neither strategy degrades VQA quality. RadGrounder thus enables spatially verifiable predictions at no cost to language performance.

\myparagraph{Limitations}
Our training data comes from a single hospital, and multi-center validation is needed to confirm generalization. The spatial grounding targets anatomical structures (via TotalSegmentator) rather than pathological findings, limiting clinical utility for lesion-level localization. Finally, comparing detection and segmentation grounding is inherently difficult, as the two approaches solve different geometric tasks (box overlap vs.\ pixel-level mask agreement), making direct comparison via a single metric nontrivial.

\myparagraph{Future Work}
Natural extensions include multi-center training to improve generalization and pathology-level grounding using lesion annotations, which would enable clinically actionable spatial predictions beyond anatomical localization.

\clearpage
\begin{credits}
\subsubsection{\ackname}
This research was funded by the
Deutsche Forschungsgemeinschaft (DFG, German Research Foundation)
417962828,
539134284,
and 499552394 -- SFB 1597 -- Small Data,
as well as through
EFRE (FEIH\_2698644) and the state of Baden-Württemberg.
\begin{center}
\includegraphics[width=0.3\textwidth]{logo_bw.png} ~~~
\includegraphics[width=0.3\textwidth]{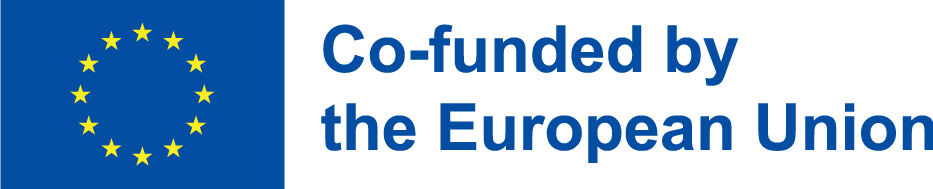}
\end{center}
\end{credits}

\bibliographystyle{splncs04}
\bibliography{ref}

\end{document}